\documentclass[letterpaper, 10pt, journal, twoside]{IEEEtran}

\usepackage{adjustbox}
\usepackage{graphicx}
\usepackage{subfig}
\usepackage[percent]{overpic}
\graphicspath{{./figs}}
\usepackage{epsfig}

\usepackage{amsmath,amssymb,amsfonts}
\usepackage{algorithm}
\usepackage{algpseudocode}
\usepackage{xcolor}
\usepackage[colorlinks=true, linkcolor=black, urlcolor=cyan, filecolor=black, citecolor=black]{hyperref}
\usepackage{url}
\usepackage{amsmath,bm}
\usepackage{multirow, makecell}
\usepackage{adjustbox}
\usepackage{comment}
\usepackage{svg}
\usepackage{tabularx}
\usepackage{multirow} 
\usepackage{subcaption}
\usepackage{booktabs}
\usepackage{makecell}
\usepackage{enumitem}
\usepackage{stfloats} 
\usepackage{tikz}
\usepackage{calc} 
\usetikzlibrary{calc}
\usepackage{pifont} 
\newcommand{\cmark}{\ding{51}} 

\newcommand{\rev}[1]{{\color{black} #1}} 

\usepackage{cite} 
\usepackage{booktabs}
\usepackage{colortbl}
\newcommand{\boldmax}[1]{\textbf{#1}}
\definecolor{colA}{gray}{0.96} 
\definecolor{colB}{gray}{0.92} 

\begin{document}

\title{


Anomaly detection for generic failure monitoring in robotic assembly, screwing and manipulation
}
\author{Niklas Grambow$^1$, Lisa-Marie Fenner$^1$, Felipe Kempkes$^1$, Philip Hotz$^1$, \\Dingyuan Wan$^1$, Jörg Krüger$^2$ and Kevin Haninger$^1$ 
\thanks{\noindent$^1$ Department of Automation at Fraunhofer IPK, Berlin, Germany. \\ $^2$ Department of Industrial Automation Technology at TU Berlin, Germany. \\ Corresponding author: {\tt niklas.grambow@ipk.fraunhofer.de}}
\thanks{This project has received funding from the European Union's Horizon 2020 research and innovation programme under grant agreement No 101058521 — CONVERGING.}}



\newcommand\copyrighttext{%
  \footnotesize \textcopyright 2026 IEEE. Personal use of this material is permitted.
  Permission from IEEE must be obtained for all other uses, in any current or future
  media, including reprinting/republishing this material for advertising or promotional
  purposes, creating new collective works, for resale or redistribution to servers or
  lists, or reuse of any copyrighted component of this work in other works. DOI: 10.1109/LRA.2026.3664647}
\newcommand\copyrightnotice{%
\begin{tikzpicture}[remember picture,overlay]
\node[anchor=south,yshift=8pt] at (current page.south) 
  {\fbox{\parbox{\dimexpr\textwidth-\fboxsep-\fboxrule\relax}{\copyrighttext}}};
\end{tikzpicture}%
}

\maketitle
\copyrightnotice

\begin{abstract}


Out-of-distribution states in robot manipulation often lead to unpredictable robot behavior or task failure, limiting success rates and increasing risk of damage. Anomaly detection (AD) can identify deviations from expected patterns in data, which can be used to trigger failsafe behaviors and recovery strategies. 
Prior work has applied data-driven AD on time series data for specific robotic tasks, however the transferability of an AD approach between different robot control strategies and task types has not been shown. Leveraging time series data, such as force/torque signals, allows to directly capture robot–environment interactions, crucial for manipulation and online failure detection. 
As robotic tasks can have widely signal characteristics and requirements, AD methods which can be applied in the same way to a wide range of tasks is needed, ideally with good data efficiency. 
We examine three industrial robotic tasks, 
robotic cabling, screwing, and sanding, each with multi-modal time series data and several anomalies. Several autoencoder-based methods are compared, and we evaluate the generalization across different robotic tasks and control methods (diffusion policy-, position-, and impedance-controlled). This allows us to validate the integration of AD in complex tasks involving tighter tolerances and variation from both the robot and its environment. 
Additionally, we evaluate data efficiency, detection latency, and task characteristics which support robust detection. The results indicate reliable detection with AUROC exceeding 0.96 in failures in the cabling and screwing task, such as incorrect or misaligned parts and obstructed targets. In the polishing task, only severe failures were reliably detected, while more subtle failure types remained undetected.   

\end{abstract}

\section{Introduction} \label{sec:intro}
As robots are applied to tasks with less structure, variation in the environment, task, or robot often leads to novel, out-of-distribution states. The behavior of the robot - both classical programs and machine learning (ML)-based - in such situations is typically unknown. 
To avoid robots acting in novel and potentially dangerous situations, anomalies can be detected \cite{Graabæk_2023_experimental_ad}, e.g. triggering a human intervention or failsafe behaviors. 
For broad deployment of AD in robotic manipulation, it is desired that the same approach can be applied to a variety of tasks and robot control strategies. Furthermore, an understanding of what properties of failures allow their detection can help predict how effective AD will be. 
Anomaly detection in robotic manipulation has been shown in specific tasks, e.g. pick and place applications, validating the ability to detect task- and robot-related deviations in force/torque signals, motor currents, and task state \cite{Graabæk_2023_experimental_ad, Brockmann_2023_vorausad}. However, \cite{Brockmann_2023_vorausad} indicate that AD in contact is less accurate, and feasibility studies as well as datasets for contact-rich robotic tasks remain limited.
\begin{figure}[t]
  \centering
  \includegraphics[width=0.48\textwidth]{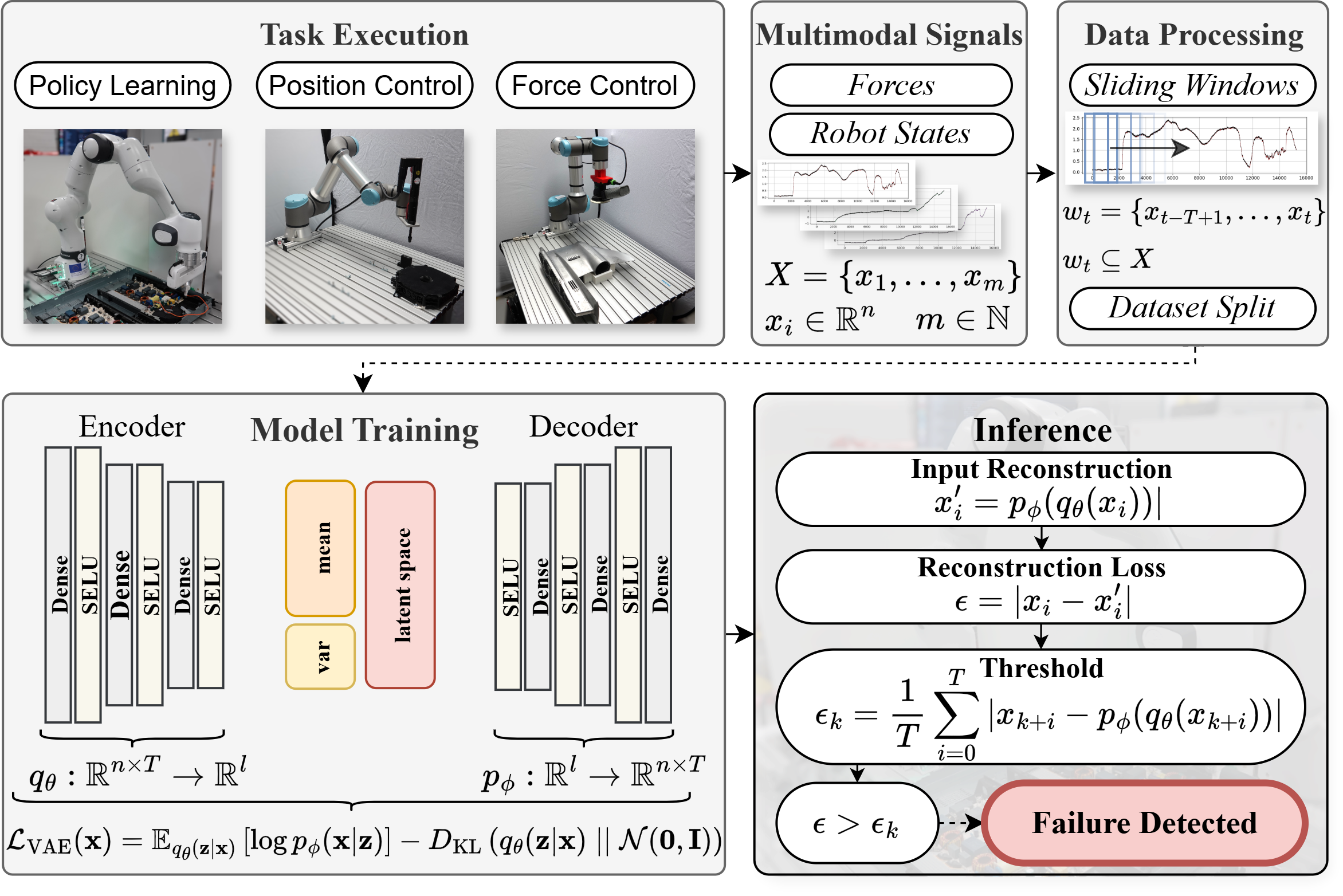}
  \caption{Principal sketch of how training and inference works for the anomaly detection including a visualization of each application.}
  \vspace{-0.5cm}
  \label{fig:framework}
\end{figure}
In contrast to previous work \cite{Chen_2020_sliding_window_online, Brockmann_2023_vorausad, Graabæk_2023_experimental_ad, Leporowski_2022_ad_dataset_screwing} that focuses solely on analyzing the general feasibility for AD in specific robotic tasks, we evaluate applicability across diverse contact-rich robotic tasks: plug insertion, screwing and polishing.
Unlike \cite{Brockmann_2023_vorausad, Graabæk_2023_experimental_ad, xu2025can}, our dataset incorporates subtle failures characterized by low-magnitude force deviations. Moreover, the dataset features multiple robot control strategies to effectively manage disturbances likely to happen in real world applications, with more variation in process time and robot trajectory.
Based on this dataset, we exclusively benchmark data-driven AD approaches including latency and factors impacting the detection performance, such as data volume or model configuration.
We demonstrate their potential for online execution, paving the way to integrate with recovery behaviors or consider within planning processes in the future.
In Section II related approaches and relevant models are introduced, then in Section III the AD framework is described including the experimental scenarios. Finally in Section IV the experimental results are presented and discussed.

\newpage
\section{Related Work}

This section reviews state-of-the-art in general anomaly detection and task-specific AD in manipulation.

\subsection{Anomaly Detection Overview}

AD identifies deviations from expected data patterns \cite{Chandola_2009_ad_survey}. These are referred to as anomalies or outliers \cite{Chalapathy_2019_deep_learning_ad_survey, Chandola_2009_ad_survey}. In robotic automation, anomalous behavior often indicates potential failures, posing risks to quality, safety, and cost - highlighting the need for robust AD.
Within AD, data-driven, model- and knowledge-based approaches can be distinguished \cite{Khalastchi_SoA}. Model-based approaches typically employ mathematical or logical descriptions of the underlying process or system, while knowledge-based methods rely on pre-established fault initializations to subsequently recognize faults both are limited in scalability, particularly as task complexity increases \cite{Chen_2020_sliding_window_online, Graabæk_2023_experimental_ad}. In contrast, data-driven approaches are referred to as model-free and solely learn the behavior from input data. Although training often requires a relatively high computational effort, the decreasing cost of computational power is making these approaches increasingly popular \cite{Graabæk_2023_experimental_ad}. Classical ML-based approaches have also been integrated for AD, including Support Vector Machine (SVM) \cite{Rodriguez_SVM} or k-Nearest Neighbors (kNN) \cite{Ando_knn}.
In semi-supervised AD, only nominal process data are required, simplifying data collection while maintaining the ability to detect unseen failures. There are two types of semi-supervised AD, forecasting and reconstruction, which differ primarily in the data used to train the models. Forecasting models predict future data points based on current observations, whereas reconstruction-based methods learn to reproduce the current data \cite{Schmidl_2022_ad_time_series_data, Darban_2022_deep_ad_time_series_survey}. Anomalies can be identified by comparing the models estimated data with the original observations, where deviations indicate abnormal behavior. Therefore, autoencoder approaches are frequently employed in various configurations due to their ability to compress data into a lower-dimensional space in the encoder and reconstruct it to its original dimensions in the decoder. A key difference between certain models is whether or not they incorporate temporal dependencies. Recurrent networks, such as Long Short-Term Memory (LSTM) architectures, are well established and have been used for both prediction \cite{Lindemann2021lstmpredict, Park2022lstmpredict} and, more commonly, for reconstruction \cite{Chen_2020_sliding_window_online, Graabæk_2023_experimental_ad, Park_2017_multimodal_ad_feeding, malhotra2016lstmrecon, cho2014lstmrecon}.

\subsection{Task-specific Approaches in Robotics}
Although many AD approaches focus on image data \cite{Inceoglu_2021_fino, yoo2021multimodal, thoduka2021visualad}, they are also applicable to time-series data in robotics, which is widely available and allows efficient computation at high sampling rates. Often, joint or end-effector positions are used in combination with other sensors, such as motor current \cite{Chen_2020_sliding_window_online}, force/torque  \cite{Graabæk_2023_experimental_ad, Brockmann_2023_vorausad}, or acoustic \cite{Park_2017_multimodal_ad_feeding} data. Ablation studies show that adding modalities generally enhances detection performance \cite{Park_2017_multimodal_ad_feeding}.
Despite the higher computational costs of ML-based approaches, the declining price of computational resources \cite{Khalastchi_SoA} has encouraged their adoption, which have potential for good scalability and efficient integration without the need for application-specific knowledge \cite{Chen_2020_sliding_window_online}. Autoencoder-based frameworks are often proposed, such as variational autoencoder (VAE) models with convolutional \cite{Chen_2020_sliding_window_online} or recurrent layers \cite{Park_2017_multimodal_ad_feeding}.    
Alternatively, some well-established methods have been applied for AD in robotics, such as Principal Component Analysis \cite{Hornung_2014_model_free_robot_ad}, Gaussian Mixture Models \cite{Romeres_2019_ADwithGMM}, Hidden Markov Models \cite{Park_2016_multimodal_ad_monitoring, azzalini2020hmm}, as well as distance-based approaches such as k-NN or Local Outlier Factor \cite{Graabæk_2023_experimental_ad}. These approaches often require extensive process knowledge that scales poorly with task complexity. In contrast, ML methods learn from demonstrations, enabling easier integration and transfer across complex manipulation tasks.

While AD has been applied to robot planning and perception tasks without physical contact \cite{sinha2024real, Hornung_2014_model_free_robot_ad}, many robotic tasks inherently involve contact \cite{Graabæk_2023_experimental_ad, Brockmann_2023_vorausad, Leporowski_2022_ad_dataset_screwing, Park_2017_multimodal_ad_feeding, Romeres_2019_ADwithGMM, Miquel_2023_tactile_sensing_peginhole, Inceoglu_2021_fino}. Therefore, AD has been validated in more general manipulation tasks, such as robot-assisted feeding \cite{Park_2017_multimodal_ad_feeding}, handling household items \cite{Inceoglu_2021_fino, yoo2021multimodal}, pick-and-place \cite{Brockmann_2023_vorausad, Graabæk_2023_experimental_ad} and peg-in-hole tasks \cite{Miquel_2023_tactile_sensing_peginhole}. As industry-relevant tasks, plug insertion \cite{Romeres_2019_ADwithGMM} and screwing \cite{Leporowski_2022_ad_dataset_screwing, Aronson2016datadrivenco, Cheng2018screwdrivingad} have been investigated.

\subsection{Failure Types and Characteristics}
We categorize the analyzed failures into two main classes: process-related failures and external disturbances. The latter failures can be prevented avoiding unintended external influences, e.g. by isolating the robot's workspace through additional protective components. Many of the proposed failure types arise from external disturbances, including events such as manual interference by an operator (e.g., physical impact to the robot) \cite{Chen_2020_sliding_window_online, Park_2017_multimodal_ad_feeding, Graabæk_2023_experimental_ad}, as well as occlusions \cite{Park_2017_multimodal_ad_feeding}, interference from nearby objects \cite{Graabæk_2023_experimental_ad} or vibrations. Process-related failures, in contrast, arise from the robotic system or task dynamics and typically occur during repeated task execution, mostly independent of the specific experimental setup. These include phenomena such as grasp slippage \cite{yoo2021multimodal, Brockmann_2023_vorausad, Graabæk_2023_experimental_ad}, incorrect localization \cite{Inceoglu_2021_fino}, or failures caused by the intrinsic characteristics of the manipulation process, e.g. incorrect insertion due to tight tolerances \cite{Miquel_2023_tactile_sensing_peginhole} or variations in the target object \cite{Leporowski_2022_ad_dataset_screwing, Graabæk_2023_experimental_ad}. Many proposed failure types are caused by external disturbances that lead to major signal deviations and lack of industrial relevance. In contrast, process-related failures (e.g., misaligned plug) typically produce subtler changes. Our dataset includes a range of failures, varying in cause and severity, from minor anomalies with slight signal deviations to severe cases with significant signal changes.      

\subsection{Robot Control and Datasets}
Deploying AD works in parallel with a robot program which is responsible for the task execution. The robot's execution can use various control strategies which have different characteristics such as the trajectory variance. Position-controlled approaches have minimal variation in motion trajectories, reducing variance in signals for the AD \cite{yoo2021multimodal, Leporowski_2022_ad_dataset_screwing, Romeres_2019_ADwithGMM}. More complex control strategies leverage image \cite{Brockmann_2023_vorausad} and force \cite{ Graabæk_2023_experimental_ad} information to locate the object to be manipulated, assess the target position or ensure compliance \cite{Park_2017_multimodal_ad_feeding}. Advanced control has higher variation in the robot execution, and more variation in nominal task performance. Additionally, AD based on tactile information can be used to detect changes in the task and switch controller \cite{Miquel_2023_tactile_sensing_peginhole}. 
The existing datasets \cite{Graabæk_2023_experimental_ad, Brockmann_2023_vorausad, Leporowski_2022_ad_dataset_screwing}  do not yet cover realistic industrial robot tasks, particularly those involving rich contact and force-sensitivity, nor do they compare the ability of AD methods to generalize across hardware platform or task. Although \cite{Graabæk_2023_experimental_ad} and \cite{Brockmann_2023_vorausad} utilized image and contact information to estimate the target position, AD has not been evaluated in learning-based control setups, which may have different characteristics regarding variance. High task execution variance in imitation (IL) and end-to-end learning can interfere with considering temporal relations, as proposed in \cite{Park_2017_multimodal_ad_feeding, Graabæk_2023_experimental_ad, Brockmann_2023_vorausad}.

\section{Anomaly Detection Approach}

This section outlines the proposed AD approach, covering data processing, training, inference, and evaluation metrics.

\subsection{Data Processing} \rev{The time series data is described by $X = \{ \mathbf{x}_{1}, \mathbf{x}_{2}, \dots, \mathbf{x}_{t_{m}}\}$ with the number of time steps $t_{m} \in \mathbb{N}$ and $\mathbf{x}_{t}\in \mathbb{R}^{n}$ is an $n$-dimensional data point representing the concatenated input modalities. The data from the task execution is processed using sliding windows $\mathbf{w_{t}}$ with specified window length $T$. The windows advance by a single time step to achieve maximal overlap, preserving high temporal fidelity in the dataset. The dataset consists of a trajectory from each run, and each trajectory is windowed independently to avoid data leakage between runs. A sequence $\mathbf{w}_{t}$ at time step $t$ is composed of $\mathbf{w}_{t} = \{\mathbf{x}_{t-T+1}, \dots, \mathbf{x}_{t}\} \subseteq X$ \cite{Chen_2020_sliding_window_online}. Data normalization by mean/variance was tested, where no performance gains were observed, therefore unnormalized data were used.} 

\subsection{Architecture} Autoencoder architectures comprise an encoder and a decoder, where the decoder architecture is mirrored from the encoder. This study considers different autoencoder models used for AD, outlined in the following section.

\subsubsection{Variational Autoencoder (VAE)}
The architecture of the VAE model is illustrated in \autoref{fig:framework}. VAEs are generative models that learn a probabilistic latent representation of the input data \cite{kingma}. A combined loss function is minimized that includes both a reconstruction loss and the Kullback-Leibler (KL) divergence (see \autoref{eq:kl_loss}). The latter minimizes the learned distribution with a predefined prior, a standard normal distribution to approximate the latent representation $p(z) = \mathcal{N}(0, I)$. KL regulation encourages the model to generate the latent space according to the desired distribution \cite{nguyen2024variational}.
\subsubsection{Autoencoder (AE)}
A simple AE optimizing only reconstruction loss was evaluated, with an unchanged encoder-decoder structure (\autoref{fig:framework}).
\rev{
\subsubsection{Undercomplete Autoencoder (UAE)}
Another architecture was validated, in which the input dimension exceeds the latent dimension, following the parameters proposed in \cite{Graabæk_2023_experimental_ad}. 
\subsubsection{SWCVAE}
The architecture proposed in \cite{Chen_2020_sliding_window_online} modifies a VAE model by introducing two convolutional layers that process the input data before flattening, ensuring compatibility with the initial input shapes.
}
\subsection{Training} During training a model is built that projects time sequences from the training data into a lower dimensional latent space representation. 
\rev{
Each timestep $\mathbf{x} \in \mathbf{w}$ is reconstructed so that $\mathbf{w'} = \{\mathbf{x}_1', \dots, \mathbf{x}_T'\}$, where $T$ is the window size and $\mathbf{x'}$ is the reconstruction of a given timestep $\mathbf{x}$.  
The reconstruction error is then utilized for optimization \cite{Chen_2018_ae_theory}. 
We consider the Euclidean distance (L2-loss) for computing the reconstruction loss $\epsilon$ minimized in each training step. It can be described by
\begin{equation}
\label{eq:delta_recon}
    \epsilon(\mathbf{{w}}, \mathbf{{w}}') = \sqrt{\sum_{\mathbf{x} \in \mathbf{w}} \sum_{i=1}^{n} (\mathbf{x}(i) - \mathbf{x}'(i))^{2}},
\end{equation} 
where $i\in [1,\dots, n]$ indexes the feature.
}
Depending on the specific model architecture additionally KL-divergence loss term can be considered
\begin{equation}
\label{eq:kl_loss}
\text{D}_{\text{KL}} = \sum_{j=1}^{n} \left( 1 + \log(\sigma_{j}(q_{\theta})^2) - \mu_{j}(q_{\theta})^2 - \sigma_{j}(q_{\theta})^2 \right).
\end{equation}

After training on nominal tasks, the reconstruction loss increases for anomalous states. 

\subsection{Threshold Computation}
\label{subsec:threshold}

Due to the absence of temporal relationships between individual windows, we propose a novel approach in which thresholds are derived from the reconstruction of the nominal process data. \rev{The trained model is then used to reconstruct each window $\mathbf{w}$ from unseen nominal samples $\mathbf{{w}'} = p_{\phi}\!\left(q_{\theta}(\mathbf{w})\right)$, where $q_{\theta}$ and $p_{\phi}$ denote the encoder and decoder functions, respectively. For each of the windows, the mean absolute reconstruction error is computed feature-wise:
\begin{equation}
L_{i} (\mathbf{w})= \frac{1}{T}\sum_{\mathbf{x} \in \mathbf{w}}|\mathbf{x}(i)-\mathbf{{x}}(i)'|, 
\end{equation}
where $T$ is the window length and $i\in [1,\dots, n]$ indexes the feature. 
Subsequently, for each feature, a distinct threshold $\epsilon_i = \max_{\mathbf{w} \in W} L_i(\mathbf{w})$ is determined by identifying the maximum reconstruction error observed across all windows $W$, thereby defining the upper limit of expected behavior under normal conditions including task variance. }To mitigate false positives, a  tolerance factor may be added to the threshold; however, this may cost sensitivity to actual failures. Once trained, the model can be deployed for online AD. \rev{In the experimental validation, a dedicated model is trained for each task, and feature-wise thresholds are likewise derived independently for each task. For online validation, a combined loss score was introduced by applying per-feature min–max scaling.}

\subsection{Evaluation Metrics}

Following prior work, we evaluate AD using standard metrics proposed in \cite{Graabæk_2023_experimental_ad, Brockmann_2023_vorausad, Chen_2020_sliding_window_online, Park_2017_multimodal_ad_feeding}, summarized in \autoref{tab:metrics}. Combined with industrial metrics (e.g., data efficiency and detection latency), these evaluations help quantify performance, reliability, and commissioning effort for AD across tasks.

\begin{table}[b]
\vspace{0.3cm}
\centering

\renewcommand{\arraystretch}{1.1}
\footnotesize
\begin{tabular}{@{}p{2.8cm}p{5.4cm}@{}}
\toprule
\textbf{Metric} & \textbf{Definition} \\
\midrule
\multicolumn{2}{l}{\textit{Detection Metrics}} \\
Precision & \( \text{TP}\cdot(\text{TP}+\text{FP})^{-1} \): correct positives \\
Recall (TPR) & \( \text{TP}\cdot(\text{TP}+\text{FN})^{-1} \): detected positives \\
F1-score & \( 2PR\cdot(P+R)^{-1} \): harmonic mean of P = Precision, R = Recall \\
FPR & \( \text{FP}\cdot(\text{FP}+\text{TN})^{-1} \): false positives ratio \\
ROC & TPR vs. FPR curve across thresholds \\
AUROC & Area under ROC (higher is better) \\
\midrule
\multicolumn{2}{l}{\textit{Industry Relevant Metrics}} \\
Data Efficiency & AUROC vs. training data volume \\
Detection Time & Avg. latency to detect failure \\
\bottomrule
\end{tabular}
\vspace{1mm}
\begin{minipage}{0.9\linewidth}
\vspace{0.1cm}
\footnotesize
TP / FP = True / False Positives,\quad TN / FN = True / False Negatives 
\end{minipage}
\caption{Metrics used for evaluation.}
\label{tab:metrics}
\end{table}

\section{Experimental Evaluation}

\begin{figure*}[b]
    \centering
    \begin{minipage}{0.32\textwidth}
        \centering
        \includegraphics[width=\linewidth]{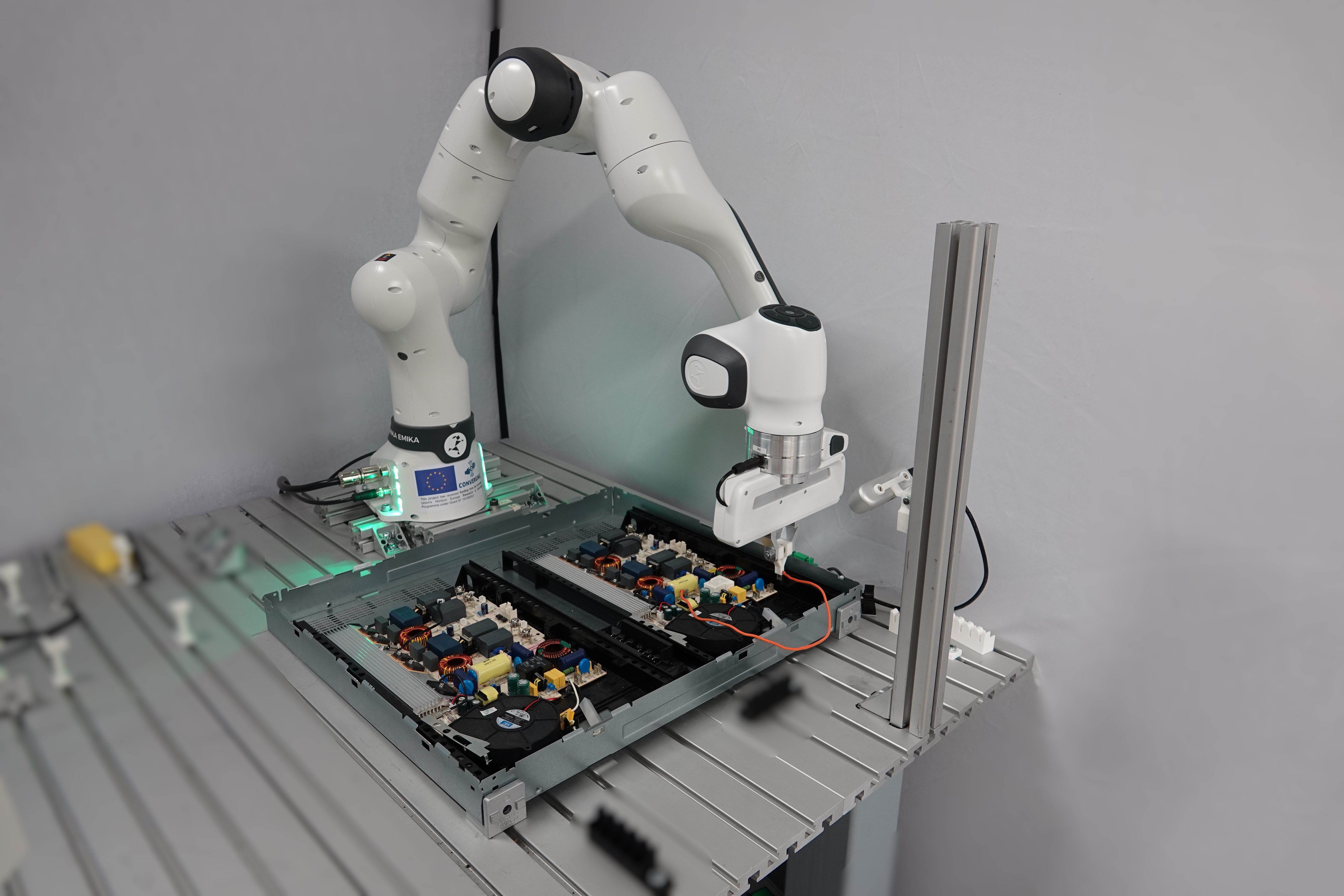}
        \caption*{1) Cabling Task}
    \end{minipage}
    \hspace{0.005\textwidth}
    \begin{minipage}{0.32\textwidth}
        \centering
        \includegraphics[width=\linewidth]{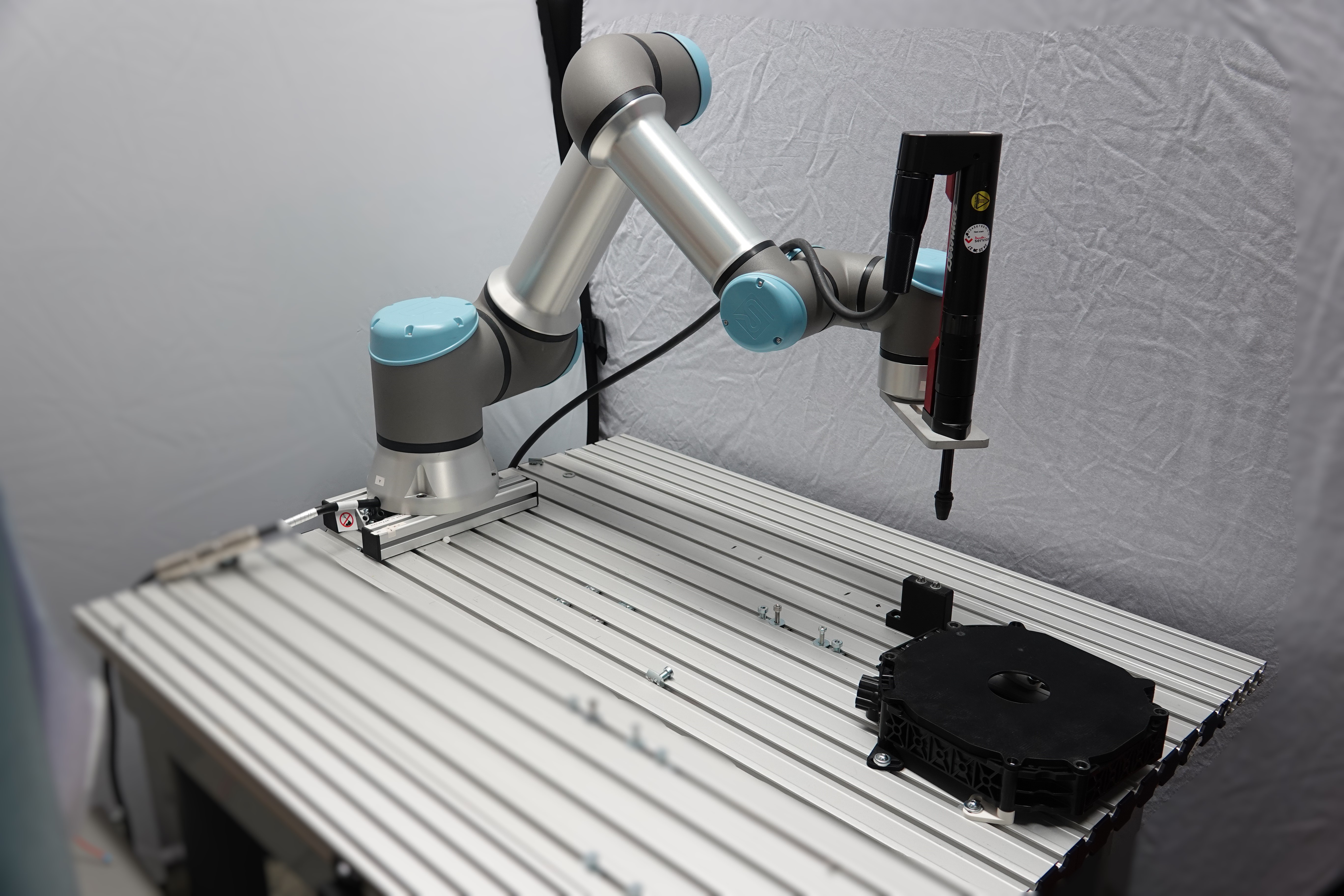}
        \caption*{2) Screwing Task}
    \end{minipage}
    \hspace{0.005\textwidth}
    \begin{minipage}{0.32\textwidth}
        \centering
        \includegraphics[width=\linewidth]{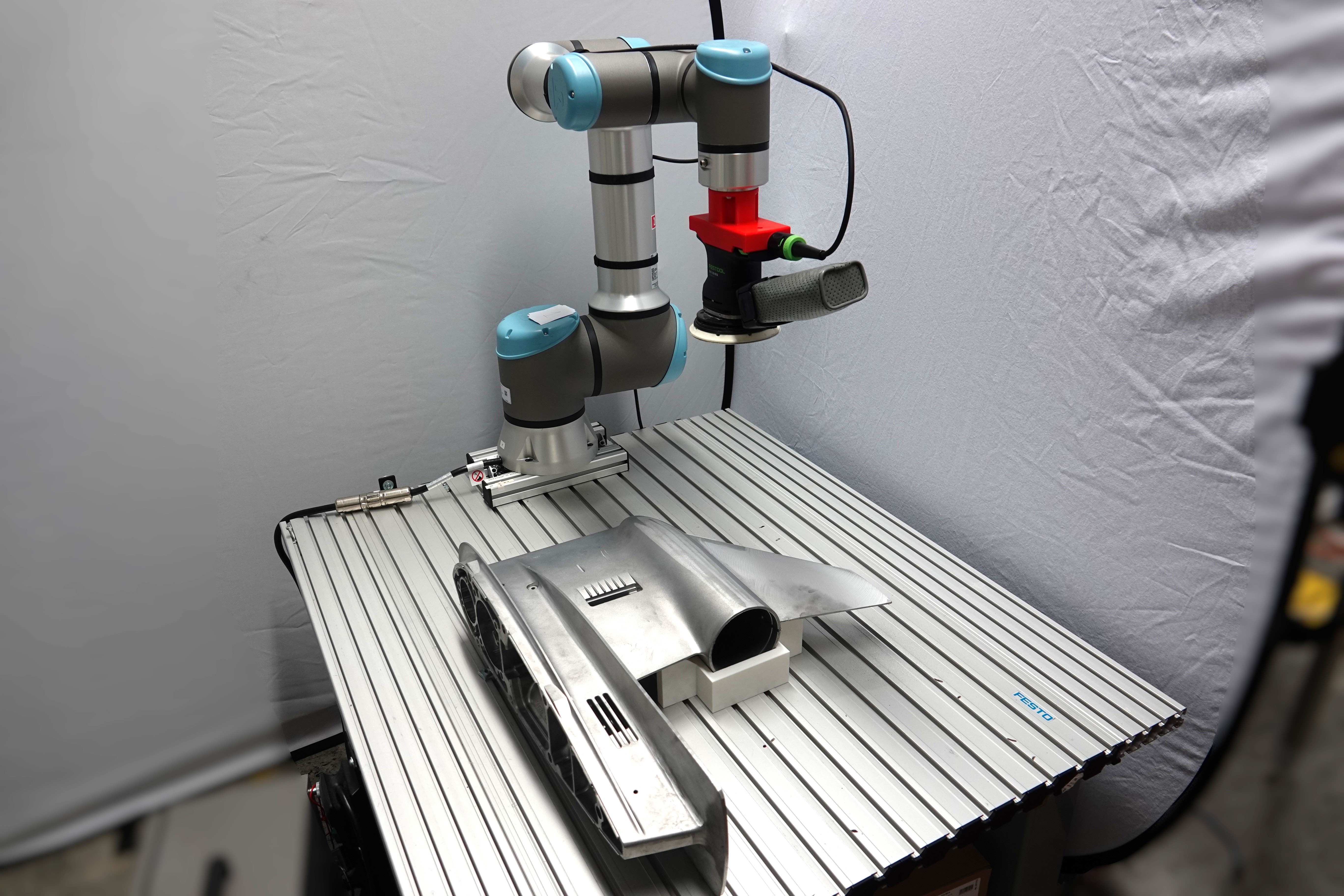}
        \caption*{3) Polishing Task}
    \end{minipage}

    \caption{Experimental setups for various industrial applications including plug insertion, screwing and polishing.}
    \label{fig:exp_setup}
\end{figure*}

Within the experimental evaluation, we propose three tasks involving contact, plug insertion, screwing, and polishing, as illustrated in \autoref{fig:exp_setup}.  

\subsection{Experimental Setup}

\subsubsection{Cabling}
The contact task is derived from an industrial use case in electric stove assembly. The robot must first grasp a female plug and then insert it into a vertically oriented male connector on the stove surface. An IL approach is employed for robot control \cite{chi2024diffusionpolicy}, using camera inputs alongside TCP pose as input. Expert demonstrations are collected using a SpaceMouse, allowing for precise and intuitive control of the robot’s movements-especially during the critical phase of approaching the connector. During inference the controller reaches up to $5$\:Hz cycle time. The demonstrations, originally gathered for policy learning, are reused to train the AD model.

\subsubsection{Screwing}
The industrial use case involves the assembly of vehicle battery housing parts using multiple screws. The robot must first pick up a screw and place it onto the threaded surface of the housing at a specified position. It then proceeds to insert and tighten the screw. Position control was employed for robot operation, based on distinct reference points defined beforehand. In this setup, additional nominal task executions are required to gather data for training the AD model.  

\subsubsection{Polishing}
The industrial process involves the polishing of aluminum casting parts. A manual polishing tool, equipped with sandpaper, is mounted on the robot’s end-effector. Using a Cartesian force controller \cite{FDCC}, the robot follows a predefined trajectory while maintaining a desired contact force along the polishing direction. The goal is to polish specific surface regions until the target surface quality is achieved, characterized by a lower roughness than the initial state. 
The AD model, trained on polished-part data, guides the robot on rough parts until reconstruction error drops below a threshold, indicating task success. Defining anomalous states inversely allows monitoring during polishing to prevent damage and enable targeted post-processing. A secondary upper threshold is introduced to safely halt operations and prevent failures like over-polishing, where reconstruction loss rises after reaching the target state.

\subsection{Collected Dataset}
The dataset comprises both nominal and failure scenarios, summarized in \autoref{tab:dataset}. For each application, the robot's wrench forces/torques, joint positions, velocities, and gripper positions (if available) are recorded. To improve AD evaluation, temporal labels are introduced for failure scenarios, provided by an operator at the moment an anomaly occurs.

\begin{table}[b]
    \centering
    \footnotesize
    \renewcommand{\arraystretch}{1.2}
    \begin{tabular}{c|l|c|c|c}
        \toprule
        \multicolumn{2}{c|}{\textbf{Application}} & \textbf{1) Cabling} & \textbf{2) Screwing} & \textbf{3) Polishing} \\
        \midrule
        \multirow{3}{*}{\rotatebox[origin=c]{90}{\textbf{Parameter}}}
        & Robot   & \makecell{Franka\\R. 3} & \multicolumn{2}{c}{UR16e} \\
        & Axes    & 7 & \multicolumn{2}{c}{6} \\
        & Control & \makecell{Diff.\\ Policy\cite{chi2024diffusionpolicy}} & \makecell{Position\\Control} & \makecell{Force\\Control \cite{FDCC}} \\
        \midrule
        \multirow{4}{*}{\rotatebox[origin=c]{90}{\textbf{Dataset}}} 
        & \# Nominal Runs       & 20 & 140 & 30 \\
        & $\mu$ Duration [s] & 43.1 & 17.2 & 39.0 \\
        & $\sigma^2$ Duration [s] & 3.4 & 0.1 & 1.1 \\
        & \# Anomaly Runs     & 50 & 40 & 50 \\
        & Frequency [Hz] & \multicolumn{3}{c}{500} \\
        \bottomrule
    \end{tabular}
    \caption{Dataset parameters for industrial applications. Average duration and variance are computed across runs; frequency denotes the sampling rate.}
    \label{tab:dataset}
\end{table}

\subsection{Failure Cases}
Selected failure scenarios reflect realistic challenges in industrial process development, covering variations in forces, signals, contact modes, and failure duration.

\subsubsection*{\textbf{1) Cabling}}

\begin{enumerate}[label=\alph*), leftmargin=4em]
\item \textbf{Human Disturbance:} External force is applied to the robot's wrist in motion, simulating a collision.
\item \textbf{Misaligned Plug:} A rotated plug stands upright instead of sliding in, causing insertion failure and potential damage.
\item \textbf{Incorrect Target Pose:} Workpiece displacement alters the end effector target, often causing plug misalignment and mechanical stress.
\item \textbf{Different Connector:} A smaller, incompatible connector prevents full insertion due to size mismatch.
\item \textbf{Missing Plug:} No cable or wrong object (e.g., pen) is grasped, simulating setup errors.
\end{enumerate}

\subsubsection*{\textbf{2) Screwing}}

\begin{enumerate}[label=\alph*), leftmargin=4em]
\item \textbf{Missing Screw:} Screw is absent at the pickup location, preventing assembly.
\item \textbf{Obstructed Screw Hole:} Screw insertion is blocked.
\item \textbf{Misalignment:} Screwdriver misses the hole and contacts the casing.
\item \textbf{Missing Thread:} Lack of threading results in altered contact and failed assembly.
\end{enumerate}

\subsubsection*{\textbf{3) Polishing}}

\begin{enumerate}[label=\alph*), leftmargin=4em]
\item \textbf{Human Disturbance:} Robot motion is disrupted by external force at the robot's wrist during polishing.
\item \textbf{Different Surface State:} Rough areas are treated as anomalies.
\item \textbf{Wrong Sanding Paper:} Incorrect grit level used.
\item \textbf{Displaced Object:} Object displacement less than 3\:cm leads to inadequate polishing.
\item \textbf{Simulated Defect:} A rigid obstacle simulates an unpolished surface requiring rework.
\end{enumerate}

\begin{figure*}[b]
    \centering
    \begin{tikzpicture}
        \node[inner sep=0pt] (plots) at (0,0) {
            \begin{minipage}{\widthof{\includegraphics[width=0.32\textwidth]{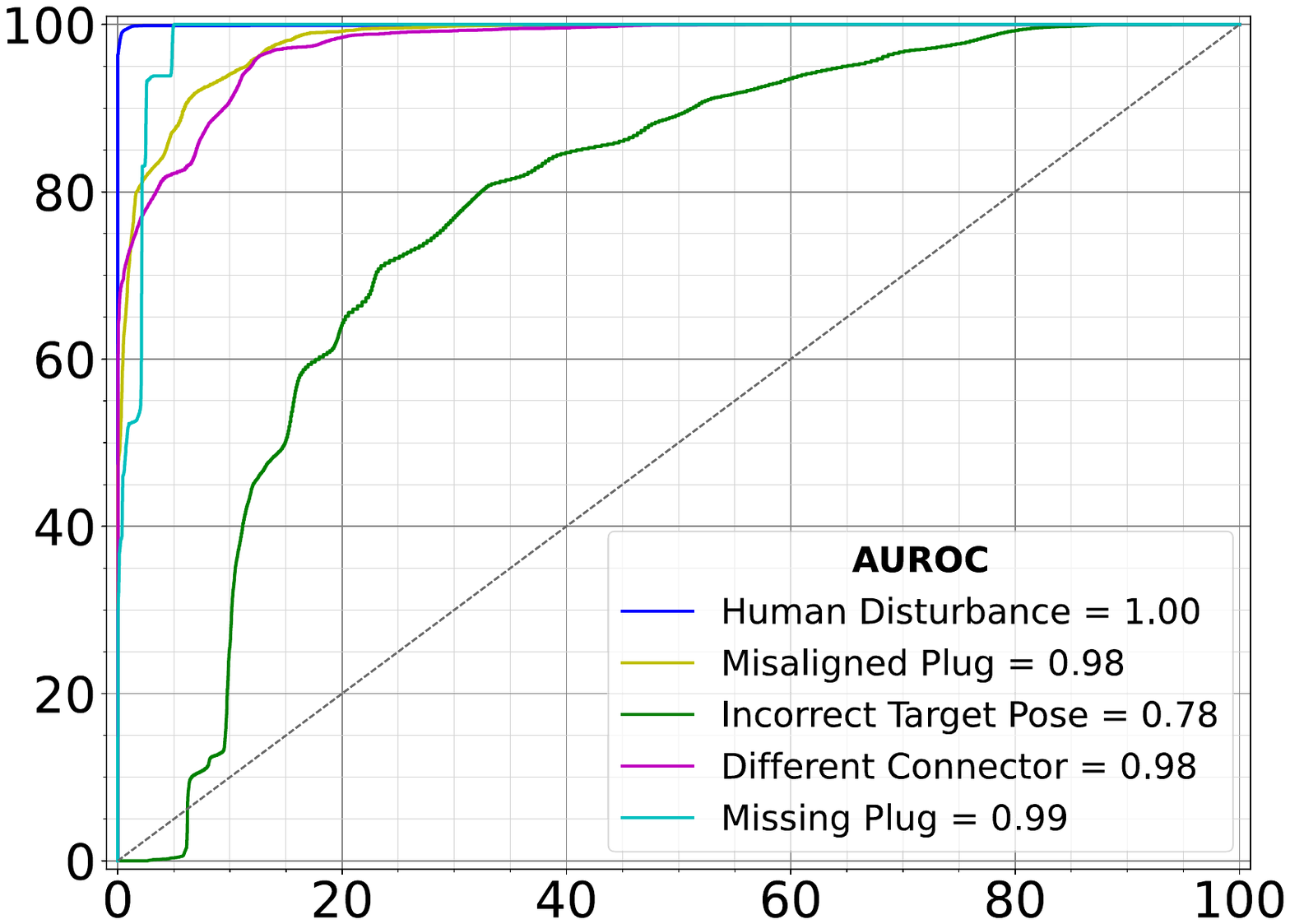}} * 3 + 0.02\textwidth}
                \centering
                \begin{minipage}{0.32\textwidth}
                    \centering
                    \includegraphics[width=\linewidth]{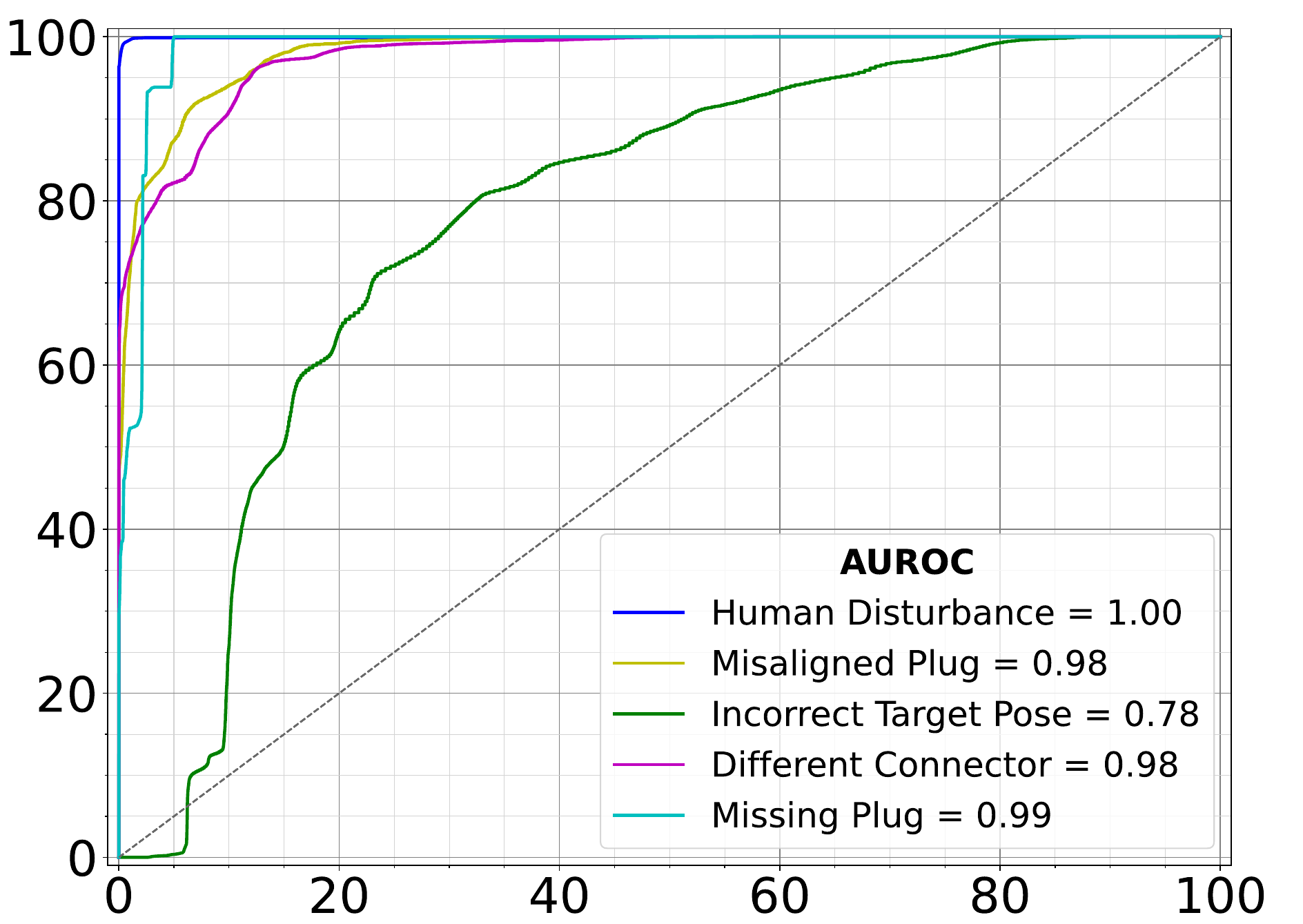}
                \end{minipage}
                \begin{minipage}{0.32\textwidth}
                    \centering
                    \includegraphics[width=\linewidth]{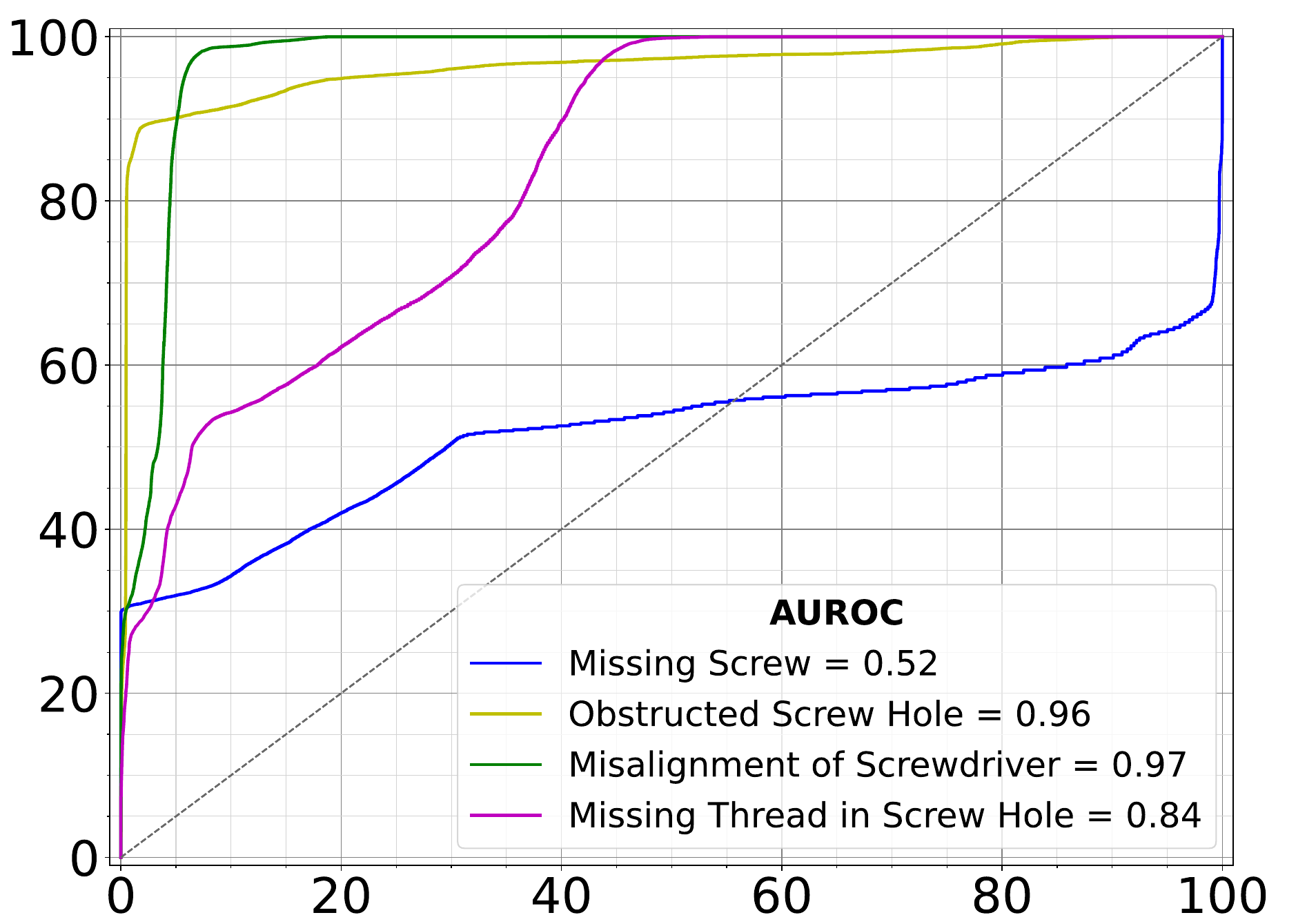}
                \end{minipage}
                \begin{minipage}{0.32\textwidth}
                    \centering
                    \includegraphics[width=\linewidth]{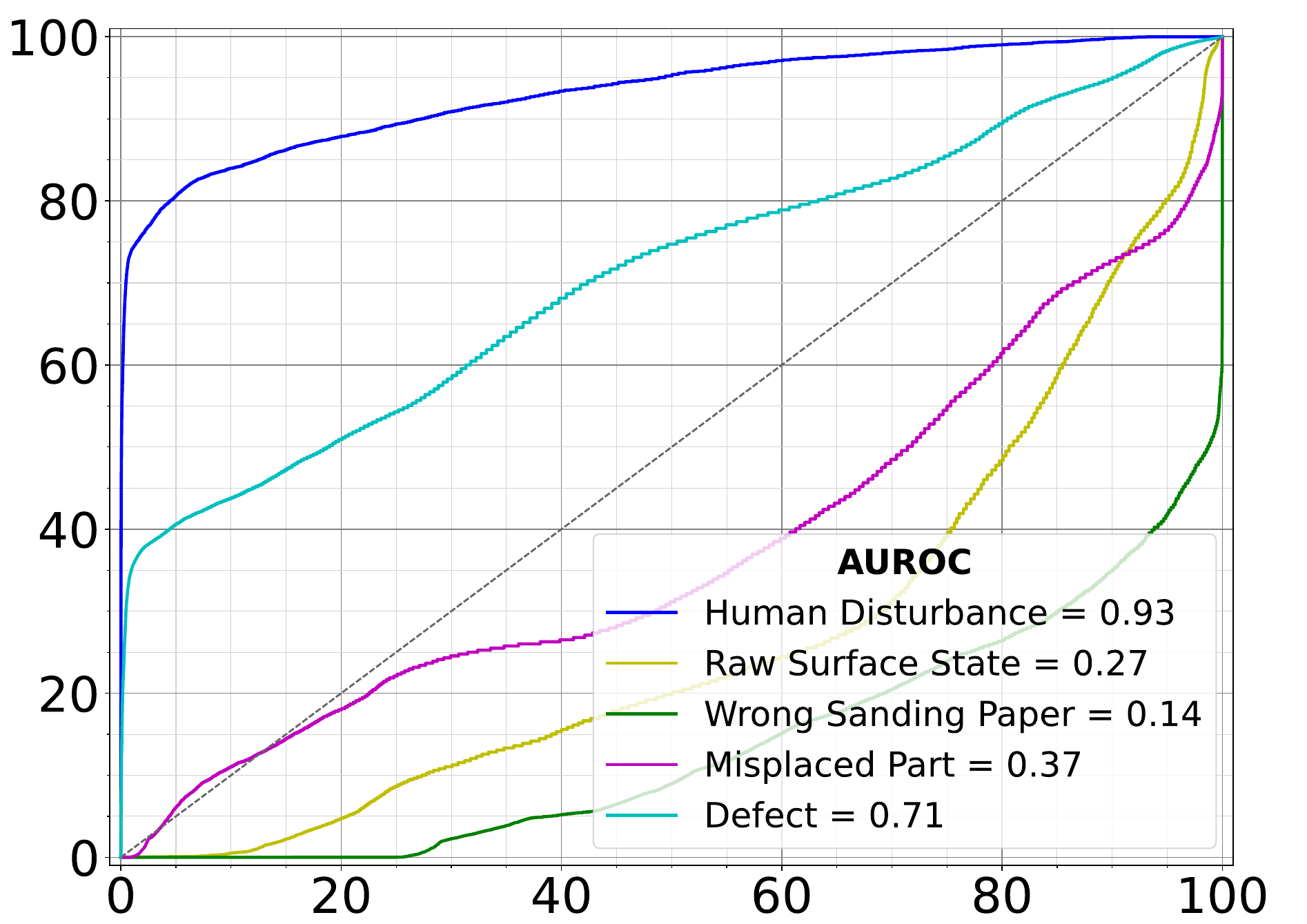}
                \end{minipage}
            \end{minipage}
        };
        \node[rotate=90, anchor=center] at ($(plots.west) + (-0.15, 0)$) {\footnotesize True Positive Rate (TPR) [\%]};
        \node[anchor=north] at ($(plots.south) + (0, -0.1)$) {\footnotesize False Positive Rate (FPR) [\%]};
        \node at ($(plots.south west) + (0.155\textwidth, -1.0)$) {\footnotesize 1) Cabling Task};
        \node at ($(plots.south) + (0, -1.0)$) {\footnotesize 2) Screwing Task};
        \node at ($(plots.south east) + (-0.155\textwidth, -1.0)$) {\footnotesize 3) Polishing Task};
    \end{tikzpicture}
    \caption{\rev{ROC curves for different applications to evaluate the VAE model's performance in each failure case across all applications.}}
    \label{fig:roc}
\end{figure*}

\subsection{Implementation details}
For model initialization, a preliminary hyperparameter search yielded the following configuration: window size $T=100$, batch size 25, latent dimension 15, and 500 training epochs. The encoder comprises three fully connected layers with 30 units each (see \autoref{fig:framework}), using SELU activation functions as introduced in \cite{klambauer2017self}, followed by a latent bottleneck. The decoder mirrors this architecture.
The proposed approach is implemented in Python 3.11 using JAX \cite{jax2018github} as ML framework. For experimental deployment, Robot Operating System (ROS) is used - specifically, ROS Noetic and ROS 2 Humble \cite{doi:10.1126/scirobotics.abm6074}. To acquire experimental data, relevant ROS topics are recorded using rosbags. During inference, a ROS node initializes the trained model and processes incoming sensor data in real time using sliding windows. The robot’s motion is immediately stopped upon detection of an anomaly. The specific inputs used for each task during training, inference and validation (\autoref{fig:roc}) are shown in Table \ref{tab:input_signals}. \rev{During data collection, individual time steps ($\mathbf{x}_{t}$) are manually labeled as anomalous to support model evaluation. During validation, sliding windows ($\mathbf{w}_{t}$) of the data are processed and classified as anomalous if any value within the window exceeds the predefined feature-wise anomaly threshold. As a result, the occurrence of a single anomalous time step is sufficient to cause the entire window to be labeled as anomalous.} 
The code repository is available at \url{https://gitlab.cc-asp.fraunhofer.de/ipk_aut/techmodules/encoder_dynamics}.

\begin{table}[h!]
\centering
\label{tab:input_signals}
\setlength{\tabcolsep}{2.5pt}
\renewcommand{\arraystretch}{0.95}
\begin{tabular}{l|c|c|c|c}
\toprule
\textbf{Signal / Task} & \textbf{Dim} & \textbf{1) Cabling} & \textbf{2) Screwing} & \textbf{3) Polishing} \\
\midrule
EE Pos., TCP ($x, y, z$)                            & 3 & \cmark & \cmark & \cmark \\
Force, TCP ($F_{x}, F_{y}, F_{z}$)               & 3 & \cmark & \cmark & \cmark \\
Vel., TCP ($v_{x}, v_{y}, v_{z}$)                & 3 & \cmark & \cmark & \cmark \\
Gripper Pos.\ ($l, r$)                           & 2 & \cmark & -       & -        \\
\bottomrule
\end{tabular}
\caption{\rev{Input signals per task. \cmark denotes measured signals; Dim indicates signal dimensionality; TCP specifies the reference frame (tool center point).}}
\end{table}

\section{Experimental Results}

The following section presents and discusses the results from the experimental validation.

\subsection{Generalizability Towards Various Applications}

In \autoref{fig:roc}, the detection performance for different failure types is evaluated across the tasks using ROC curves. Four out of five failure types in the cabling task were reliably detected. Human Disturbance and Missing Plug achieved near-perfect detection (Area under ROC curve, AUROC above 0.99). Misaligned and Incorrect Plug types, both altering contact conditions, were detected with AUROC $>$ 0.98. In contrast, the Incorrect Target Pose failure was more difficult to detect, with a significantly lower AUROC of 0.78. This comparably poor performance motivates an ablation study (see \autoref{fig:ablation}) focused specifically on this failure type.
\rev{For Screwing, two out of four failure types (Misaligned Screwdriver and Obstructed Screw Hole) were detected reliably, each achieving an AUROC above 0.96. A Missing Thread was detected with lower AUROC of 0.84.} In contrast, the Missing Screw failure proved more challenging, with the model performing close to random guessing with AUROC 0.52.
\rev{For Polishing, Human Disturbance achieves the highest AUROC 0.93 followed by the Defect with 0.71. The differences in Surface Roughness (AUROC up to 0.27) and Misplaced Part were not reliably detectable. The poor detection performance during polishing errors b), c) and d) can be explained by the high force variance in the nominal task relative to the difference in force values in the anomaly case (see Table IV, failure characteristics).} 

\begin{table*}[htbp]
\vspace{0.25cm}
\centering
\resizebox{\textwidth}{!}{
\begin{tabular}{
    ll
    |>{\columncolor{colA}}c >{\columncolor{colB}}c >{\columncolor{colA}}c >{\columncolor{colB}}c >{\columncolor{colA}}c >{\columncolor{colB}}c
    |>{\columncolor{colA}}c >{\columncolor{colB}}c >{\columncolor{colA}}c >{\columncolor{colB}}c >{\columncolor{colA}}c
    |>{\columncolor{colB}}c >{\columncolor{colA}}c >{\columncolor{colB}}c >{\columncolor{colA}}c >{\columncolor{colB}}c >{\columncolor{colA}}c
}
\toprule

\rowcolor{white}
\multicolumn{2}{c|}{\textbf{Application}} &
\multicolumn{6}{c|}{\textbf{1) Cabling}} &
\multicolumn{5}{c|}{\textbf{2) Screwing}} &
\multicolumn{6}{c}{\textbf{3) Polishing}} \\

\rowcolor{white}
\multicolumn{2}{c|}{\textbf{Failure Scenario}} &
\cellcolor{colA} a) & \cellcolor{colB} b) & \cellcolor{colA} c) & \cellcolor{colB} d) & \cellcolor{colA} e) & \cellcolor{colB} OK &
\cellcolor{colA} a) & \cellcolor{colB} b) & \cellcolor{colA} c) & \cellcolor{colB} d) & \cellcolor{colA} OK &
\cellcolor{colB} a) & \cellcolor{colA} b) & \cellcolor{colB} c) & \cellcolor{colA} d) & \cellcolor{colB} e) & \cellcolor{colA} OK \\
\midrule

\multirow{2}{*}{\makecell[l]{Failure \\ Characteristics}} 
& Force Diff. [N]           
& 8 & 11 & 6 & 11 & 0.2 & 7
& 27 & 31 & 32 & 33 & 37
& 36 & 28 & 32 & 28 & 31 & 28 \\

& Variance [$\mathrm{N}^2$]            
& 8 & 3 & 2 & 7 & 0 & 3
& 23 & 26 & 20 & 29 & 28
& 29 & 28 & 36 & 28 & 25 & 28 \\
\midrule

\multirow{12}{*}{\makecell[l]{Model \\ Performance}} 

& \multirow{2}{*}{\makecell[l]{F1-score (ours)}} 
& 54 & 72 & 21 & 71 & \boldmax{89} & --
& \boldmax{77} & 67 & 58 & 53 & --
& 39 & 38 & 44 & 33 & \boldmax{56} & -- \\
& & $\pm$ 4.7 & $\pm$ 0.8 & $\pm$ 4.4 & $\pm$ 2.8 & $\pm$ 13.9 & --
& $\pm$0.0 & $\pm$0.0 & $\pm$0.0 & $\pm$0.0 & --
& $\pm$1.8 & $\pm$1.4 & $\pm$1.1 & $\pm$2.8 & $\pm$4.6 & -- \\

\addlinespace[0.3em]  

& \multirow{2}{*}{\makecell[l]{F1-score (max)}} 
& 99 & 75 & 52 & 81 & \boldmax{100} & --
& 88 & \boldmax{93} & 87 & 68 & --
& 82 & 93 & 93 & \boldmax{95} & 57 & -- \\
& & $\pm$ 0.0 & $\pm$ 0.6 & $\pm$ 2.7 & $\pm$1.6 & $\pm$13.9 & --
& $\pm$0.0 & $\pm$0.0 & $\pm$0.0 & $\pm$0.0 & --
& $\pm$1.1 & $\pm$0.0 & $\pm$0.0 & $\pm$0.0 & $\pm$1.9 & -- \\

\addlinespace[0.3em]

& \multirow{2}{*}{\makecell[l]{F1-score (all failures)}} 
& 39 & 24 & \boldmax{47} & 26 & \boldmax{51} & --
& \boldmax{87} & 57 & 53 & 47 & --
& 39 & 38 & 44 & 33 & \boldmax{56} & -- \\
& & $\pm$ 1.0 & $\pm$ 1.0 & $\pm$ 0.9 & $\pm$ 1.8 & $\pm$ 13.9 & --
& $\pm$0.0 & $\pm$0.0 & $\pm$0.0 & $\pm$0.0 & --
& $\pm$1.8 & $\pm$1.4 & $\pm$1.1 & $\pm$2.8 & $\pm$4.6 & -- \\

\addlinespace[0.3em]

& \multirow{2}{*}{\makecell[l]{AUROC [\%]}} 
& \boldmax{100} & 98 & 78 & 98 & 99 & --
& 52 & 96 & \boldmax{97} & 84 & --
& \boldmax{93} & 27 & 14 & 37 & 71 & -- \\
& & $\pm$ 0.0 & $\pm$ 0.1 & $\pm$ 2.7 & $\pm$ 0.1 & $\pm$ 0.0 & --
& $\pm$0.2 & $\pm$0.0 & $\pm$0.0  & $\pm$0.1  & --
& $\pm$0.1 & $\pm$0.3 & $\pm$0.2 & $\pm$1.5 & $\pm$2.6 & -- \\

\addlinespace[0.3em]

& Avg. Latency [ms]         
& 3 & 31 & 180 & 53 & 0 & --
& 0 & 39 & 2 & 92 & --
& 182 & 0 & 0 & 0 & 6 & --\\

\bottomrule
\end{tabular}}
\caption{\rev{Failure-specific evaluation of the VAE model’s anomaly detection performance. The nominal task executions for each application are abbreviated as OK. Force Difference (Diff.) is the average force range per execution for each failure scenario, while Variance denotes the mean variance per execution.}}
\label{tab:detection_sensitivity}
\end{table*}

\begin{figure*}[b]
    \centering
    \begin{minipage}{0.4\textwidth}
        \centering
        \begin{overpic}[width=7.5cm]{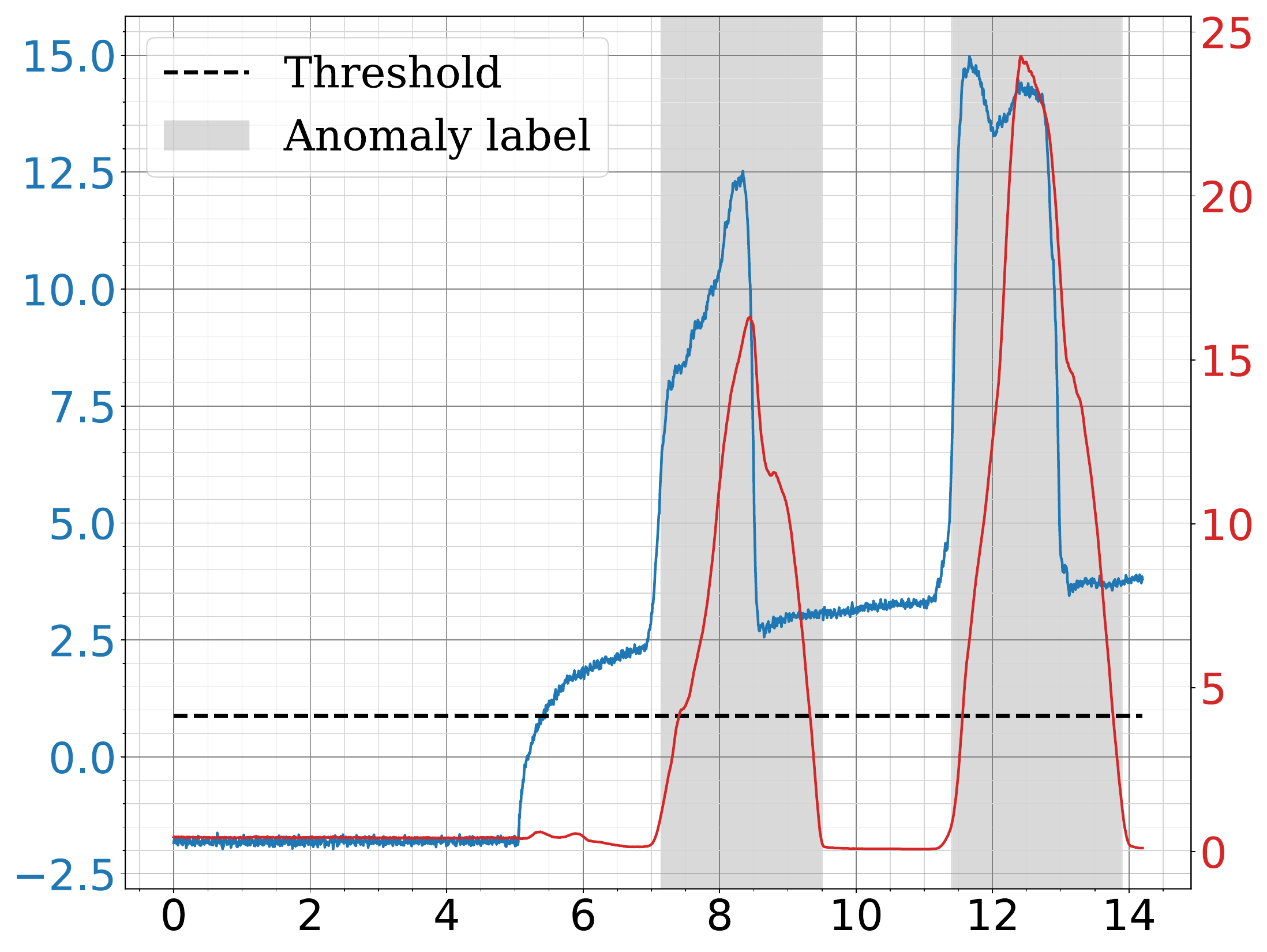}
             \put(43,-3){\footnotesize{Time [s] }}        
             \put(-5,28){\rotatebox{90}{\footnotesize{Force [N] }}}  
             \put(102,22){\rotatebox{90}{\footnotesize{Avg. RMSE [N] }}}  
        \end{overpic}
         \vspace{-6pt}
        \caption*{Exemplary visualization of the combined force reconstructions (blue), corresponding loss (red) and operator-annotated anomaly highlighted for reference (grey) for Human Disturbance (1a). }
    \end{minipage}
    \hspace{0.1\textwidth}
    \begin{minipage}{0.45\textwidth}
        \centering
        \begin{overpic}[width=7.5cm]{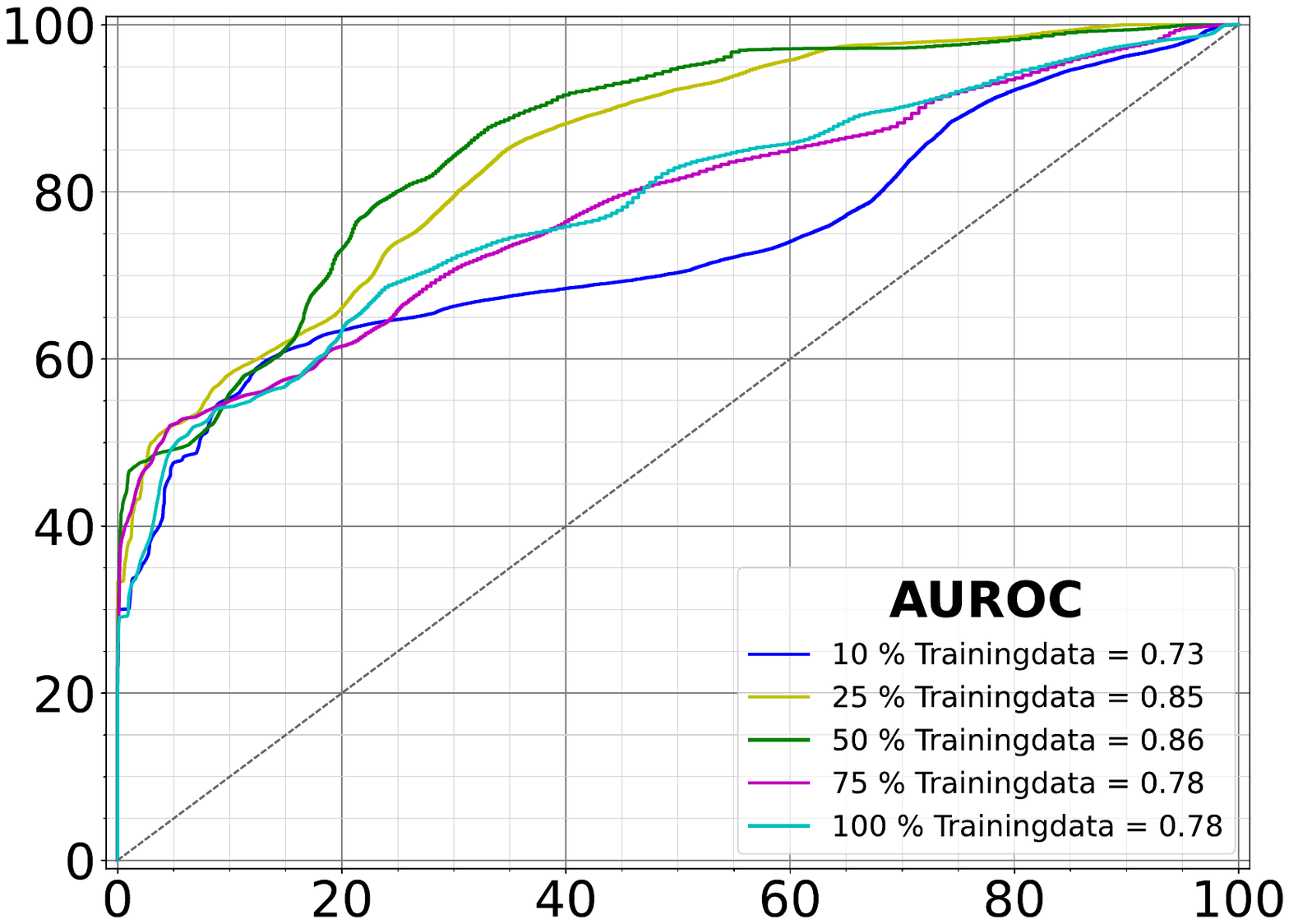}
             \put(20,-6){\footnotesize{False Positive Rate (FPR) [\%] }}        
             \put(-3,10){\rotatebox{90}{\footnotesize{True Positive Rate (TPR) [\%] }}}  
        \end{overpic}
        \vspace{14pt}
        \caption*{ROC curves for the Cabling Task, generated using models trained on varying amounts of demonstration data (2, 5, 10, 15, and 20 demonstrations), and evaluated across all failure types.}
         \vspace{-2pt}
    \end{minipage}
    \caption{Visualization of the reconstruction loss for an exemplary failure case, along with an evaluation of data efficiency both for the cabling task.}
    \label{fig:detection_latency}
\end{figure*}

\subsection{Failure Specific Performance}
To quantitatively assess the model's performance across different failure scenarios, \autoref{tab:detection_sensitivity} summarizes the F1-, AUROC, and average detection latency by failure type. Additionally each failure type is characterized by evaluating force amplitudes and variance. Beyond AUROC, threshold-specific F1-scores are also assessed to better align with industrial application needs. Since no temporal context exists between individual windows, we introduce a novel thresholding method (see \autoref{subsec:threshold}) not requiring any prior failure data. For comparison, we evaluate thresholds optimized per failure scenario and globally per tasks using the acquired failure data. The results show that selecting an appropriate threshold is challenging, e.g., in the Cabling task, where F1-scores remain low despite AUROC exceeding 0.98. Per-failure threshold optimization yields the best results. Notably, our method, performs similar to the optimized threshold across the entire failure data. For polishing, optimizing the threshold per failure improves detection, even when overall AUROC remains low. 

\subsection{Impact of Data Volume, Model Parameters and Architecture on Detection Performance}
\begin{figure*}[htbp]
    \centering

    \begin{tikzpicture}
        \node[inner sep=0pt] (plots) at (0,0) {
            \begin{minipage}{\widthof{\includegraphics[width=0.31\textwidth]{img_ROC_3/ROC_cabling_crop_2.pdf}} * 3 + 0.02\textwidth}
                \centering
                \begin{minipage}{0.31\textwidth}
                    \centering
                    \includegraphics[width=\linewidth]{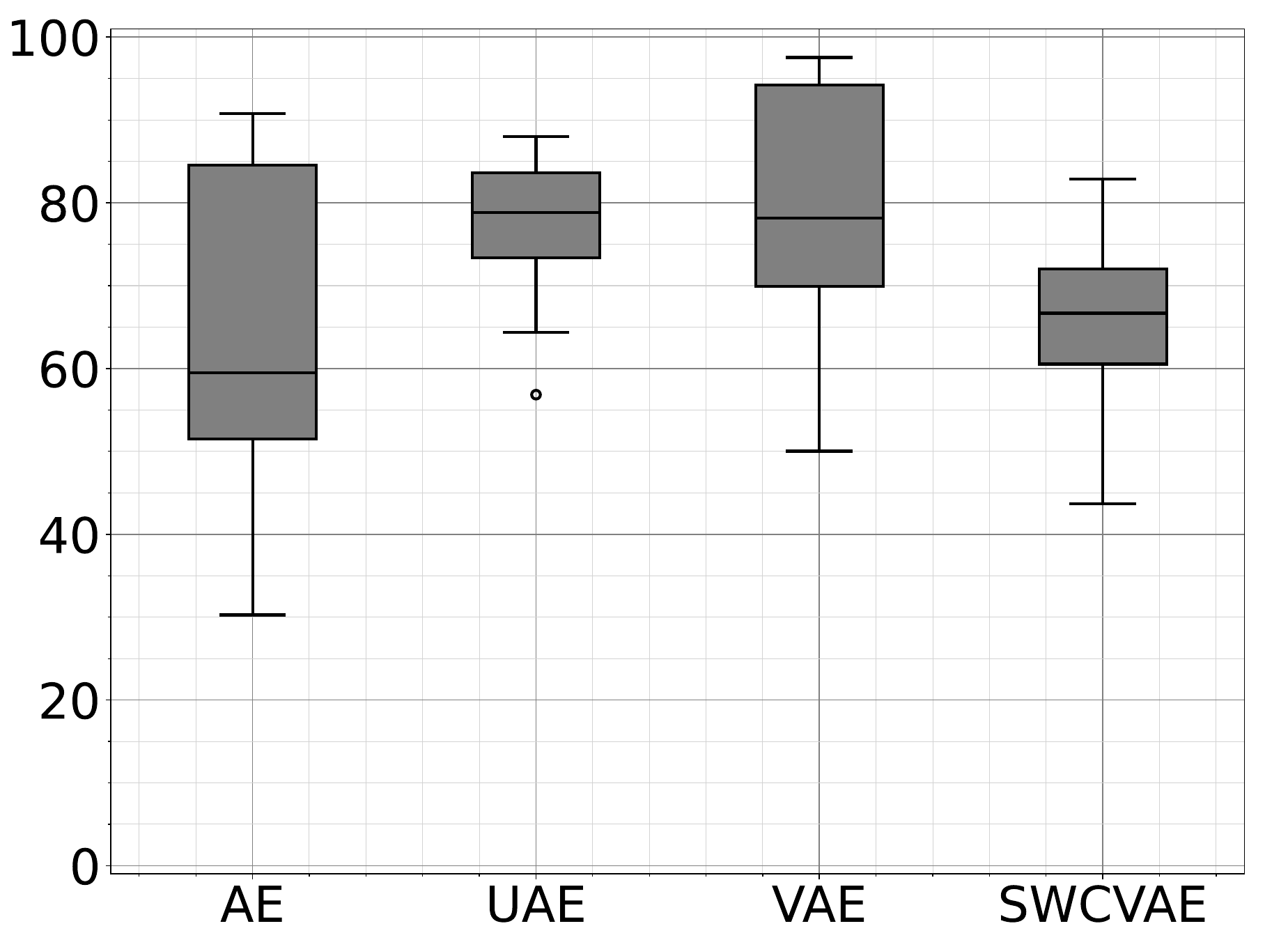}
                \end{minipage}
                    \hspace{0.01\textwidth}
                \begin{minipage}{0.31\textwidth}
                    \centering
                    \includegraphics[width=\linewidth]{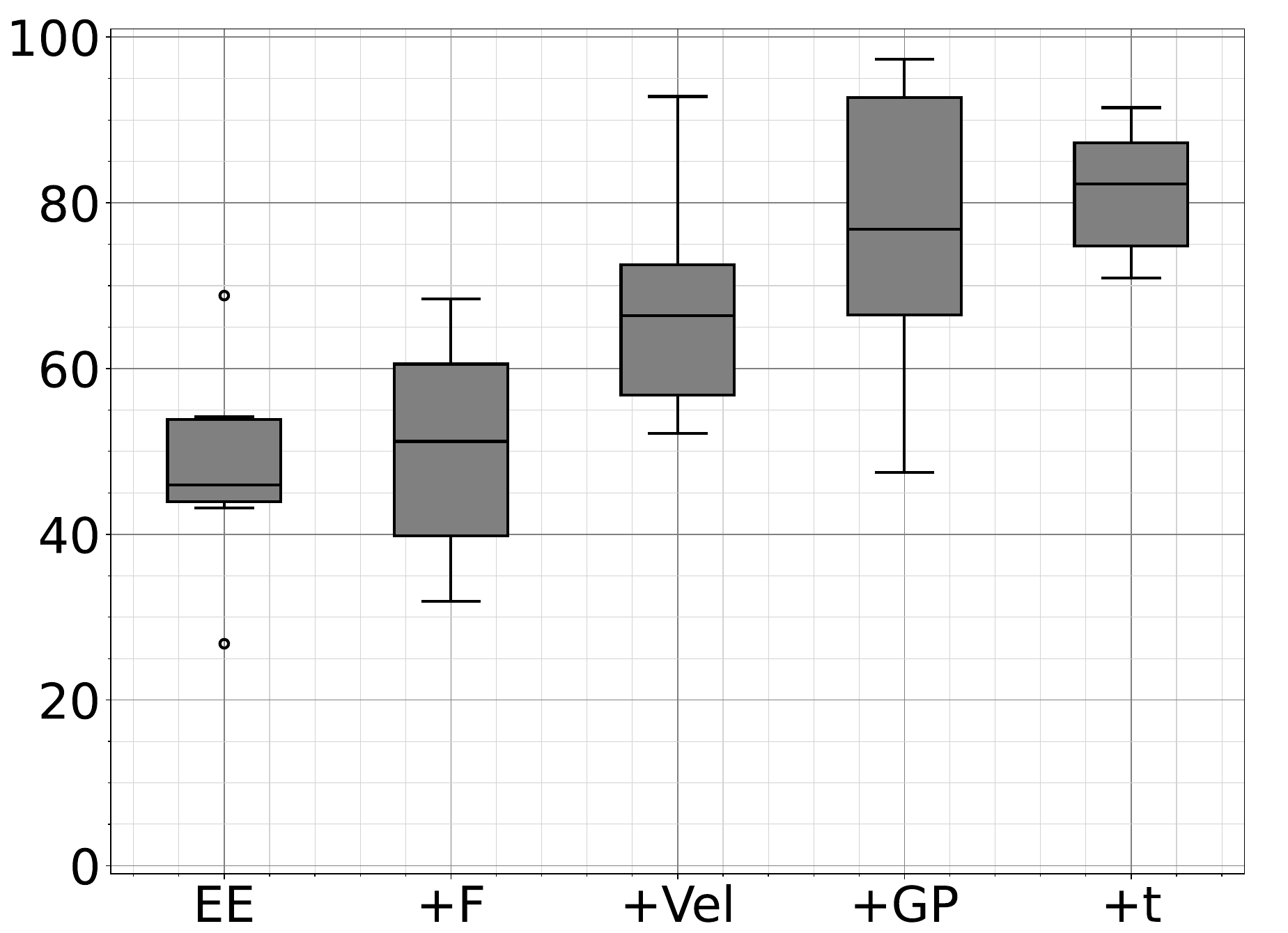}
                \end{minipage}
                    \hspace{0.01\textwidth}
                \begin{minipage}{0.31\textwidth}
                    \centering
                    \includegraphics[width=\linewidth]{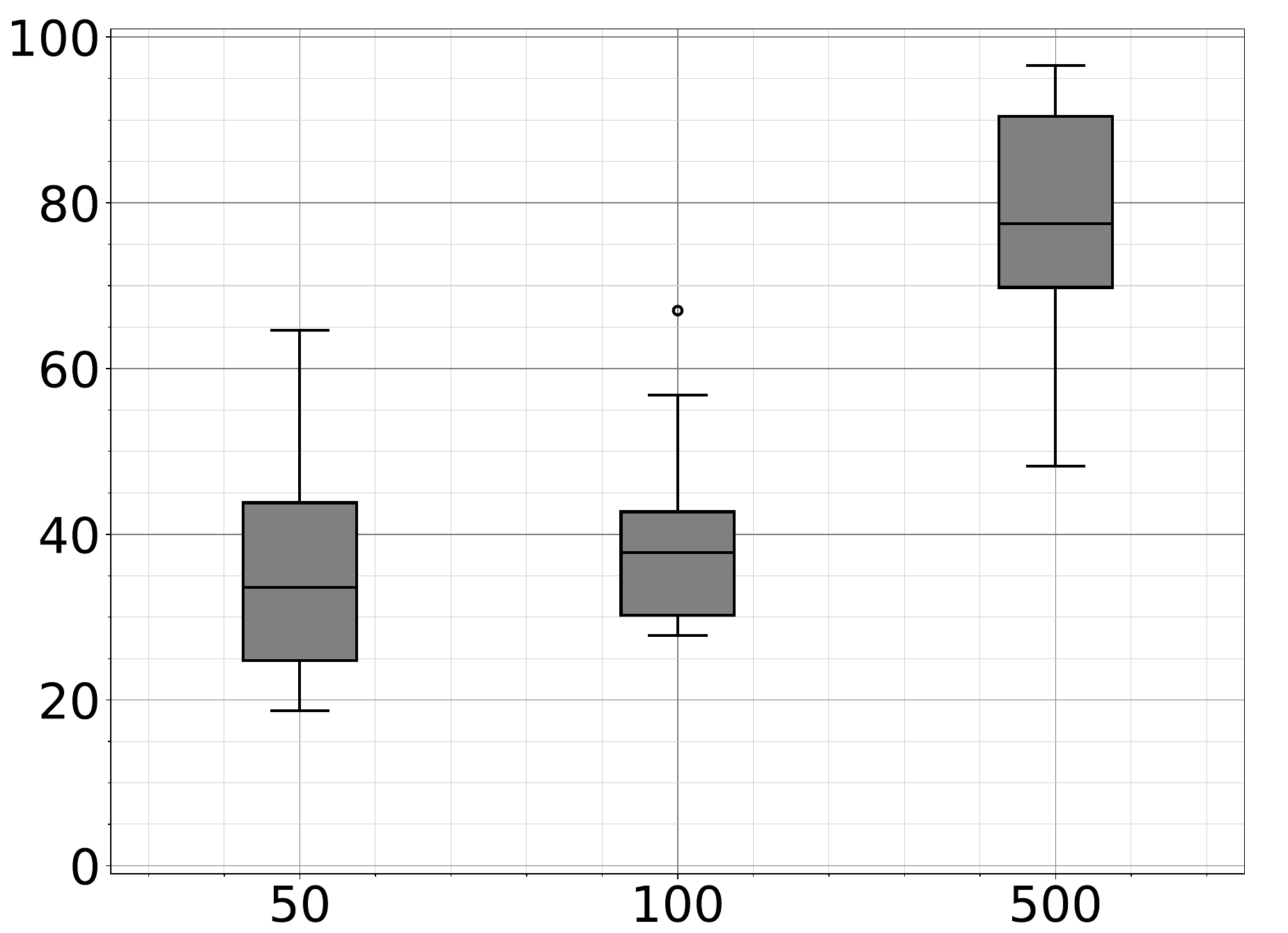}
                \end{minipage}
            \end{minipage}
        };
        \node[rotate=90, anchor=center] at ($(plots.west) + (-0.15, 0)$) {\footnotesize AUROC [\%]};
        \node at ($(plots.south west) + (0.155\textwidth, -0.3)$) {\footnotesize 1) Various Models};
        \node at ($(plots.south) + (0, -0.3)$) 
        {\footnotesize 2) Different Features};
        \node at ($(plots.south east) + (-0.155\textwidth, -0.3)$) {\footnotesize 3) Window Sizes};
    \end{tikzpicture}

    \caption{\rev{AUROC evaluation for the Cabling Task (Incorrect Target Pose Error, 1c) with varying model types, input features and window sizes. Mean and variance taken from 10 validation task executions. Evaluation with VAE, +GP, and 500 Window size unless varied in study.}}
    \label{fig:ablation}
\end{figure*}

Figure \ref{fig:detection_latency} (left) shows both the model evaluation and the operator-labeled anomaly. In this failure (1a), no contact occurs initially, with contact established around 6\:s. The reconstruction loss increases significantly at approximately 7.6\:s, coinciding visually with the labeled anomaly by the operator.
Figure \ref{fig:detection_latency} (right) evaluates the data efficiency with varying number of demonstrations from 2 to 20. The highest AUROC of 0.86 was achieved using 10 demonstrations, representing 50\% of the entire training dataset. Notably, training with the full set and 75\% of the dataset results in a slightly lower AUROC of 0.78. \rev{This trend may arise during validation, where some samples are deemed higher quality than others.} Moreover, increased variance in the demonstration data may lead to more false positives, reducing detection accuracy. Training with 5 demonstrations achieved an AUROC of 0.85, while using only 2 resulted in the lowest AUROC of 0.73. \autoref{fig:ablation} summarizes the ablation study results on Incorrect Target Pose error (1c), evaluating the impact of model architecture, feature type, and window size on detection performance. The proposed VAE model consistently outperforms both the baseline AE, which uses only dense layers, and a modified VAE variant including CNN layers (SWCVAE \cite{Chen_2020_sliding_window_online}). \rev{The undercomplete AE (UAE \cite{Graabæk_2023_experimental_ad}) produced results comparable to the VAE while exhibiting lower variance. Evaluating different modalities in \autoref{fig:ablation} (2) in addition to end-effector positions (EE), wrench force measurements (F), velocity (Vel), gripper state (GP), and the time from start of the task (t) revealed that incorporating all features yields the highest AUROC.} Additionally, window sizes of 50, 100, and 500 were evaluated. The largest window size achieved the highest AUROC (0.78). In contrast, smaller windows yielded lower mean AUROC (below 0.4), with the smallest window exhibiting notably higher variance.

\subsection{Detection Latency and Online Validation}
The detection latency, defined as the time between the labeled anomaly and the model's first tagged failure, was evaluated using optimized thresholds for each failure type. \autoref{fig:detection_latency} shows low latency (below 3\:ms) for failure (1a and 2c), despite relatively low F1-scores, likely affected by false positives. In contrast, failures (1e and 2a) show zero latency despite high F1-scores, indicating labeling inconsistencies where events were annotated prior to the actual failure.

\section{Discussion}

The proposed method, combining a VAE model with our thresholding approach, demonstrates \rev{that a single AD approach can be applied to diverse industrial applications}. Training exclusively on sensor data from nominal task execution enables efficient and scalable integration. However, the results indicate that the detection performance is highly dependent on the nature of the failure scenarios. More severe failures like human disturbances (1a and 3a), which deviate significantly from nominal execution, are easier to detect and yield higher AUROC than anomalies with smaller signal deviation, such as slight displacements. Although human disturbance failures are severe, they exhibit minor force differences due to their short duration relative to the overall measurement (see \autoref{tab:detection_sensitivity}).  
Among the evaluated tasks, cabling yielded the highest AUROC with minimal variance, in contrast to screwing and polishing, where higher signal variance in the nominal case was observed. This suggests that AD becomes more challenging as task execution variability increases.
Furthermore, the results showed highest AUROC using 50\% of the available demonstrations, suggesting effective performance with reduced training data. Incorporating additional demonstrations may increase the variance in the dataset, likely impacting detection sensitivity.
Notably, we observed no clear correlation between the underlying causes of failures and the specific features where failures were detected. For instance, in cases involving displacement issues, poor reconstruction was often observed in signals unrelated to the end-effector position. Enhancing the method's ability to pinpoint the specific modality contributing to a failure could enable more targeted and effective recovery strategies. 
\rev{Moreover, incorporating temporal context has been shown to increase AUROC and reduce variance in AUROC, as visualized in \autoref{fig:ablation} (2) and in prior work by \cite{Graabæk_2023_experimental_ad}, warranting further comparative analysis to better understand its limitations in complex tasks and control strategies.} Furthermore, we observed limitations during the online detection. In particular, sensitivity issues, e.g. variation in start pose led to false positives, posing challenges for real-time deployment. 
Selecting an appropriate threshold was challenging and significantly affected detection performance. Our initial approach required no failure data but showed limitations, such as false positives. Novel strategies that consider stage-dependent variability, e.g. higher variance during contact versus free-space motion, may be beneficial.

\section{Conclusion}
The proposed unsupervised anomaly detection framework evaluates AE models for failure detection across three industrial robotic tasks, leveraging common control strategies including policy learning, impedance- and position-control. The dataset comprises at least four distinct failure modes per application with over 330 demonstrations in total. The results demonstrate near-perfect AUROC for detecting failures with large changes in force value, such as human disturbances. More subtle failures in the cabling and screwing task, such as incorrect or misaligned parts and obstructed targets, achieved AUROC values exceeding $0.94$, indicating robust detection performance. The classification performance on the polishing task was comparatively lower, \rev{ where the nominal task has higher force variance and the anomalies smaller force deviation}. High F1 scores alongside low AUROC, likely due to under-weighted false positives, highlight the need for more suited evaluation metrics. One potential of the proposed approach lies in ensuring ML-driven control that lack robustness, particularly in out-of-distribution states where these are not tested. Reliable detecting these states and considering operators could further facilitate the integration of ML-approaches within the industry and extend robot skills.   
\bibliographystyle{IEEEtran} 
\bibliography{lib_2.bib} 
\end{document}